%% file: cvpr.tex
\documentclass[final]{cvpr}

\usepackage{times}
\usepackage{epsfig}
\usepackage{graphicx}
\usepackage{comment}
\usepackage{amsmath,amssymb} 
\usepackage{color}
\usepackage{rotating}
\usepackage{array}
\usepackage{tabu}
\usepackage{bbold}
\usepackage{appendix}
\usepackage{subfigure}
\usepackage{booktabs}
\usepackage{tabularx}
\usepackage{multirow}

\newcolumntype{L}[1]{>{\raggedright\let\newline\\\arraybackslash\hspace{0pt}}m{#1}}
\newcolumntype{C}[1]{>{\centering\let\newline\\\arraybackslash\hspace{0pt}}m{#1}}
\newcolumntype{R}{>{\raggedleft\arraybackslash}X}
\newcolumntype{P}{>{\raggedleft\arraybackslash}p{.5in}}

\newcommand*\rot{\rotatebox{90}}
\newcommand{\myparagraph}[1]{
\vspace{0.1cm}\noindent
\textbf{#1.}
}

\usepackage[pagebackref=true,breaklinks=true,colorlinks,bookmarks=false]{hyperref}

\def\cvprPaperID{14} 
\def\confYear{CVPR 2021}

\begin{document}

\title{InfoScrub: Towards Attribute Privacy by Targeted Obfuscation}

\author{Hui-Po Wang$^1$ \quad Tribhuvanesh Orekondy$^{2}$ \quad Mario Fritz$^1$\\ \\
$^1$CISPA Helmholtz Center for Information Security, Germany \\
$^2$Max Planck Institute for Informatics, Saarland Informatics Campus, Germany
}

\maketitle

\begin{abstract}

Personal photos of individuals when shared online, apart from exhibiting a myriad of memorable details, also reveals a wide range of private information and potentially entails privacy risks (e.g., online harassment, tracking).
To mitigate such risks, it is crucial to study techniques that allow individuals to limit the private information leaked in visual data.
We tackle this problem in a novel image obfuscation framework: to maximize entropy on inferences over targeted privacy attributes, while retaining image fidelity.
We approach the problem based on an encoder-decoder style architecture, with two key novelties: (a) introducing a discriminator to perform bi-directional translation simultaneously from multiple unpaired domains; (b) predicting an image interpolation which maximizes uncertainty over a target set of attributes.
We find our approach generates obfuscated images faithful to the original input images and additionally increases uncertainty by 6.2$\times$ (or up to 0.85 bits) over the non-obfuscated counterparts.



\end{abstract}

\setlength{\textfloatsep}{10pt plus 1.0pt minus 2.0pt}

\input{article/intro.tex}
\input{article/related.tex}
\input{article/method.tex}
\input{article/experiments.tex}
\input{article/conclusion.tex}

{\small
\bibliographystyle{ieee_fullname}
\bibliography{egbib}
}

\clearpage
\appendix
\input{article/appendix}

\end{document}

%% file: article/intro.tex
\section{Introduction}
\label{sec:intro}




A tremendous amount of personal visual data is shared on the internet everyday \cite{perrin2019share} e.g., camera photos shared on social networks.
The wide range of private information inadvertently leaked as a consequence is severely under-estimated \cite{orekondy17iccv}.
To prevent catastrophic side-effects of such privacy leakage (e.g., online harassment, deanonymization), it is crucial to study techniques that allow users to limit the amount of private information revealed in images before they are shared online.
To this end, we build upon recent advances in computer vision techniques and present methods to protect the privacy of individuals in visual data.

Specifically, we explore the notion of obfuscating selected privacy attributes in images.
Most literature around obfuscating image regions focus on detection and masking (e.g., by blurring) a narrow set of privacy attributes -- predominantly faces and license plates.
However, a recent line of work \cite{orekondy17iccv,orekondy18cvpr} extends these efforts to a much wider range of attributes, largely motivated that images contain various bits of information -- much like pieces of a jigsaw puzzle -- which in conjunction can compromise the individual's privacy.
In this work, we study targeted obfuscation of a variety of privacy attributes (e.g., hair color, facial hair) in images, many of which cannot be clearly localized (e.g., age).



However, as a natural side-effect, obfuscation-based approaches towards manipulating images lead to destroying the `utility' of images.
Various concepts of utility were explored recently.
The predominant notion \cite{creager2019flexibly,bertran2019adversarially,roy2019mitigating} is to define utility w.r.t a complementary set of non-sensitive utility attributes that can be inferred from images e.g., emotion.
As a result, such formulations treat obfuscation as a minimax game between inferences of disjoint privacy and utility attributes.
However, in this work, we hope to capture the usefulness of an obfuscated image beyond a (typically small) set of categorical attributes.
Consequently, our work considers the visual quality of the image as a proxy to the utility, which is inherently important for online photo sharing.

Our solution involves synthesizing images resembling the original input image, albeit with certain privacy attributes removed. 
The solution is reminiscent of a recent line of work of performing attribute manipulation on images using generative adversarial networks.
However, as we will show later, attribute-manipulation GANs pose numerous subtle problems when the task involves manipulating \textit{privacy} attributes.
In particular:
(i) they fail to associate non-removal of a particular attribute with a large (privacy) cost; 
(ii) manipulated images collapse to extreme solutions (attribute present or absent), whereas the required obfuscated solutions are typically in-between, i.e., exhibiting maximum entropy over presence/absence of the targeted attribute; and
(iii) attribute manipulations fail to generalize to unseen adversaries.
To tackle these challenges, we find existing attribute-manipulation GANs limited, and work towards attribute-\textit{obfuscation} GANs.


We present a two-stage approach towards targeted obfuscation of privacy attributes in images.
The first stage performs \textit{attribute inversion} in images: given an input image, to toggle the presence/absence of the target attribute.
We find existing image-manipulation techniques only partially invert the attributes, and hence fail to generalize to unseen adversaries.
We tackle the partial inversion problem by employing a novel bi-directional discriminator and additionally employ adversarial training to update the discriminator.
Consequently, we find the first stage of our approach more effective in inverting attributes than attribute-manipulation counterparts.
In the second stage, we extend our approach to performing \textit{attribute obfuscation} by maximizing uncertainty over the presence of the target attribute.
The key challenge here is the lack of ground-truth examples containing obfuscated images to guide the supervision.
To combat this, our second-stage model searches for the obfuscated image by interpolating the input image and the corresponding attribute-inverted image.


We evaluate our approach on CelebA by obfuscating ten facial attributes (e.g., gender, hair color), while keeping the generated image faithful to the original i.e., preserving the remaining privacy attributes and image fidelity.
We highlight that our evaluation setting is more involved than existing obfuscation literature:
(i) we consider a wider range of privacy attributes to obfuscate; and
(ii) we forego a constrained and limited set of categorical utility attributes, and solely consider the broader notion of image fidelity.
In this challenging setting, we find our approach successfully manipulate privacy attributes.
For instance, we find our approach inverts presence and absence of privacy attributes, with 84.5\% accuracy, an increase of 18.5\% achieved by recent image-translation model such as StarGAN.
Furthermore, apart from inverting privacy attributes, we find our approach equally capable of obfuscating them i.e., maximizing uncertainty of attribute predictions.
Specifically, we observe an average increase of entropy from 0.2$\pm$0.21 bits to 0.81$\pm$0.18 bits (maximum entropy = 1 bit) across inferences over ten privacy attributes.
Our results indicate we can significantly reduce the amount of private information leaked by an image, while retaining its faithfulness, and provide a viable privacy-preserving approach towards visual information sharing.

%% file: article/related.tex
\section{Related Work}
\label{sec:related_work}

\myparagraph{Attribute Manipulation}
Generative adversarial networks (GANs)~\cite{goodfellow2014generative,mirza2014conditional,odena2017conditional} have been recently extended to edit visual attributes (e.g., changing emotions in faces)~\cite{choi2018stargan,perarnau2016invertible,he2019attgan,bao2018towards} in images.
Central to these methods is using an attribute (often referred to as `domain') classifier to guide the editing process.
While these works produce often produce photo-realistic images, they are trained to fool a fixed known `adversary' (the attribute classifier).
Consequently, we find they fail to generalize to new adversaries (unseen attribute classifiers).
This is particularly problematic from a privacy stand-point, where one does not know before-hand the model used to infer attributes from images.
To tackle the generalization issue, our proposed method first trains the classifier in an adversarial manner and adopts a proposed bi-path classifier to solve the confusion problem, which is discussed in detail in Section~\ref{ssec:confused_generator}.


\myparagraph{Privacy-Preserving Learning}
In addition to attribute manipulation, several works propose to obfuscate private information from input images within a GAN-based formulation. 
To name a few, Bertran~\etal~\cite{bertran2019adversarially} learn to modify images by incorporating competition between generators and proxies of adversaries into the training, encouraging generators to better conceal sensitive information. 
Similarly, Roy~\etal~\cite{roy2019mitigating} adopts a strategy to produce privacy-preserving embeddings. 
Creager~\etal~\cite{creager2019flexibly} first learn disentangled representations with TC-VAE~\cite{chen2018isolating} and supervision from attributes. 
During the test time, it hides sensitive information by disabling the corresponding position in representations. 
While these works are effective in concealing privacy attributes, they do not generate realistic images which violates the original intention of data sharing (e.g., shared across social media). 
Moreover, except Creager~\etal~\cite{creager2019flexibly}, these algorithms need to define attributes in the training time and are not able to change privacy settings during the test time. 
In this paper, we propose a framework that provides users flexibility over a variety of sensitive attributes and obfuscate images while retaining image fidelity, which is crucial to enable photo sharing. 





%% file: article/method.tex
\section{Method}
\label{sec:method}
\begin{figure}[tbp]
    \centering
    \includegraphics[width=\linewidth]{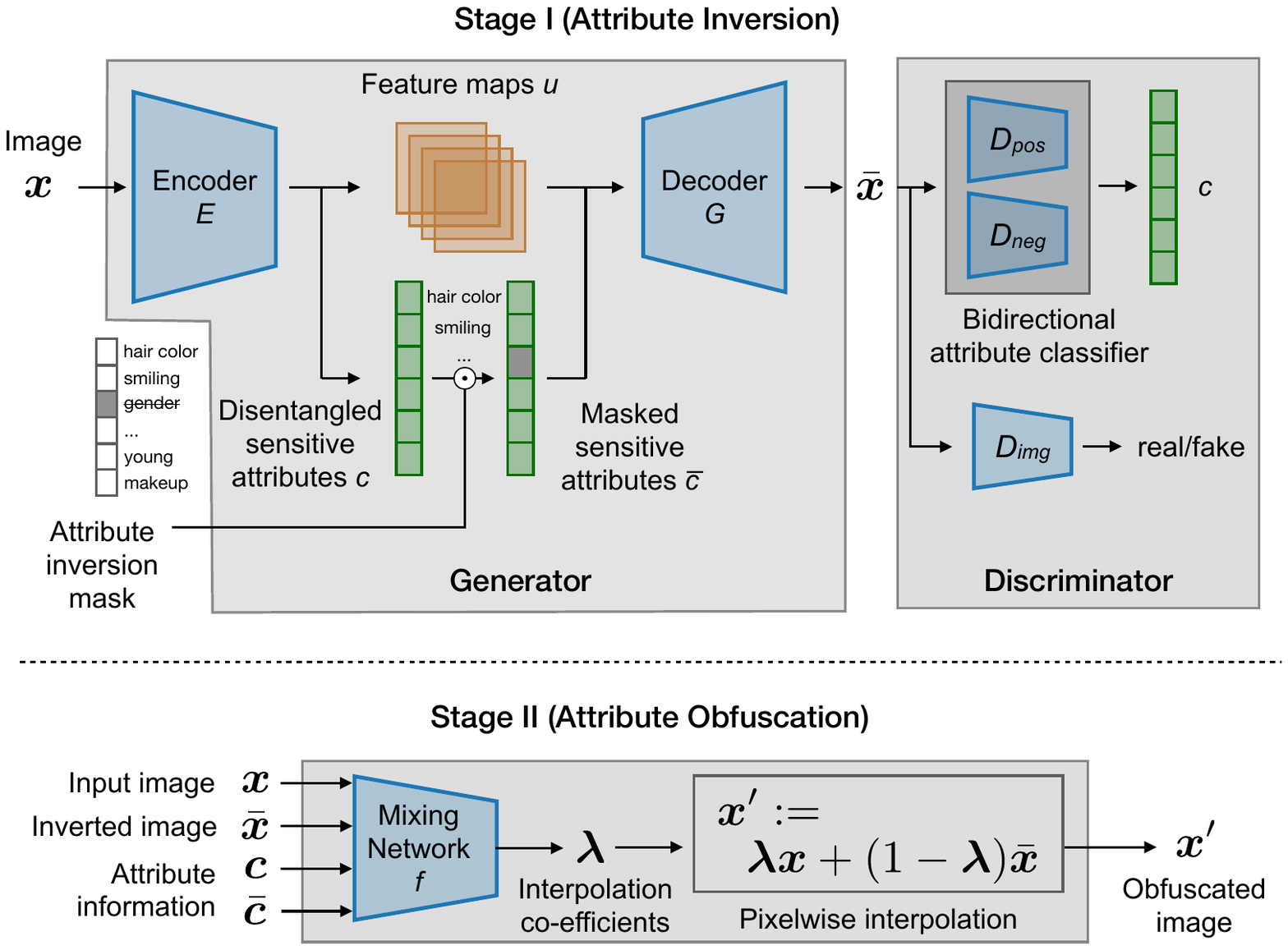}
    \caption{Our approach involves two stages: (I) we first invert the presence of the target attribute (e.g., gender) in the input image $x$ to obtain $\bar{x}$, followed by (II) crafting an obfuscated image $x'$ as an interpolation of $x$ and $\bar{x}$ to exhibit maximum uncertainty over the target attribute.}
    \label{fig:model}
\end{figure}

In this section, we present the proposed image obfuscation framework which provides users flexibility to remove an arbitrary subset of sensitive attributes while retaining the fidelity of generated images. 
Before fleshing out the details in the remainder of this section, we first provide an overview of our two stage approach (shown in Fig. \ref{fig:model}).

\myparagraph{Stage I: Attribute Inversion}
We first learn an image-to-image translation model that performs privacy attribute inversion: given an input input $x$, to synthesize an image $\bar{x}$ (faithful to $x$), but where the presence/absence of the target privacy attribute is toggled.
The key ideas in our model involve:
(i) training an encoder which disentangles the visual features in the image from the attribute information;
(ii) manipulating the disentangled attribute information to signal inversion targets; and
(iii) introducing a bi-directional discriminator, which we find crucial to alleviating issues of partial inversion of attributes.
We remark that while our approach shares some similarities with attribute manipulation strategies \cite{choi2018stargan,perarnau2016invertible}, we tackle specific challenges to better generalize to unseen attribute classifiers (critical when enforcing privacy), while producing photorealistic images.

\myparagraph{Stage II: Attribute Obfuscation}
We further extend our approach to synthesize obfuscated images i.e., images with high uncertainty over presence of target attribute.
As shown in the lower part of Fig. \ref{fig:model}, we achieve obfuscation using a mixing network, which predicts the pixel-wise linear interpolation coefficients $\lambda$ between the original input image $x$ and the attribute-inverted image $\bar{x}$.
Consequently, we arrive at an obfuscated but photorealstic image $x'$, which displays high entropy over the target attribute.

Now, we move to discussing in detail the first- (Sections \ref{ssec:method_overview}-\ref{ssec:learning}) and second-stage (Section \ref{ssec:attr_obfu}) of our approach to perform image obfuscation.




\subsection{Attribute Inversion (Stage I): Overview}
\label{ssec:method_overview}


The proposed approach transforms an input image $x \in R^{H \times W \times 3}$ to produce a complementary edited image $\bar{x}$ via an encoder-decoder architecture.
To perform arbitrary attribute inversion during a single forward-pass, the approach allows manipulation on the disentangled code produced by the encoder.
As shown in Figure~\ref{fig:model}, there are four sub-networks within the framework: an encoder $E$, a decoder $G$, a compound bi-directional attribute classifier $D_{\text{pos}}/D_{\text{neg}}$, and an image discriminator $D_{img}$.
We now discuss each of these sub-networks.


\myparagraph{Disentanglement via Encoder $E$}
To infer and decouple the underlying information, the encoder $E$ models the information by encoding input images 
\begin{equation}
\label{eq:encoding_rep}
    (u, c) = E(x),   
\end{equation}
where $u$ is a set of feature maps describing non-sensitive information, and $c$ is a vector describing sensitive attributes. The two representations are encouraged to be independent of each other.
Using the specified attribute edits (via a binary encoded inversion mask), $c$ is modified to $\bar{c}$, which is subsequently used to sanitize the input image.


\myparagraph{Decoder $G$}
With the disentangled representations, the decoder $G$ models the residuals that change the pixels with high information leakage risks. Formally, we have
\begin{equation}
\label{eq:decoding_rep}
    \bar{x} = x + G(u, \bar{c}).
\end{equation}
The motivation behind the design is that most of pixels in the input image are unrelated to sensitive information. Therefore, we can lower the cost of learning image generation by simply modeling residuals. 

\myparagraph{Image- ($D_{\text{img}}$) and Attribute-level Discriminators}
We consider two constraints on the generated image $\bar{x}$: 
(a) it should resemble realistic images; and
(b) it should fool an attribute-level discriminator trained to classify privacy attributes.
To tackle (a), we introduce an image-level discriminator $D_{img}$ to synthesize realistic images in an adversarial manner~\cite{goodfellow2014generative}.
We elaborate (b) in the next section, as naively introducing an attribute-level discriminator is problematic.

\subsection{Bi-directional Discriminator}
\label{ssec:confused_generator}

Common to many attribute manipulation techniques is employing an attribute-level discriminator (also referred to as domain classifier in literature) during training.
This discriminator (which we refer to as $D_{\text{attr}}$) is trained only on non-generated real images and is used to provide the generator $G$ feedback on whether the attribute was successfully manipulated.
Directly employing $D_{\text{attr}}$ for the problem of privacy attribute manipulation leads to two challenges.
We describe the challenges and how we address them in the following paragraphs.

\myparagraph{Discriminator Overfitting}
We find training the attribute discriminator only on real images leads to over-fitting issues, in which the model solely learns to fool the discriminator by removing specific regions of target attributes (e.g., removing the bridge of eyeglasses to eliminate the activation). 
This increases the risk that sensitive information can be still recognized from the processed images and violates our goal of protecting visual privacy. 
To alleviate the problem, one common approach is utilizing adversarial training by updating the discriminator with generated images.


\begin{figure}[t]
    \centering
    \includegraphics[width=0.95\linewidth]{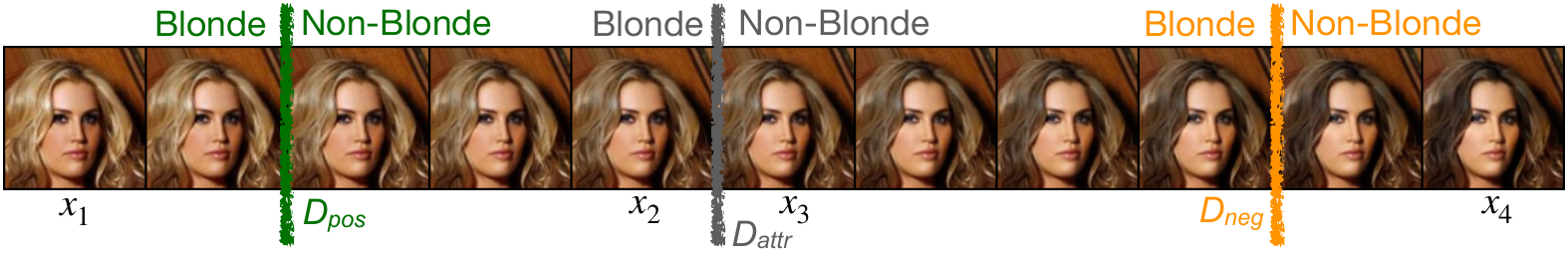} \\
    \caption{Attribute discriminator $D_{\text{attr}}$ vs. our bi-directional discriminator $D_{\text{pos}} \cup D_{\text{neg}}$. Vertical lines illustrate decision boundaries between presence and absence of attribute `blonde'. We find translating an input image (e.g., removing `blonde' attribute in $x_1$) using only $D_{\text{attr}}$ incorrectly encourages partially attribute-inverted images drawn close to the decision boundary ($x_3$). However, our discriminator learns a tighter decision boundary ($D_{\text{neg}}$) for this translation and produces an image ($x_4$) better representing inversion of the attribute. We find a similarly effective translation in the other direction as well (e.g., adding `blonde' to $x_4$ producing $x_1$) using $D_{\text{pos}}$.}
    \label{fig:conf_example}
\end{figure}



\begin{figure}[t]
\setlength\tabcolsep{0.18em}
\centering
\begin{tabular}{ccc}
 \includegraphics[width=0.32\linewidth]{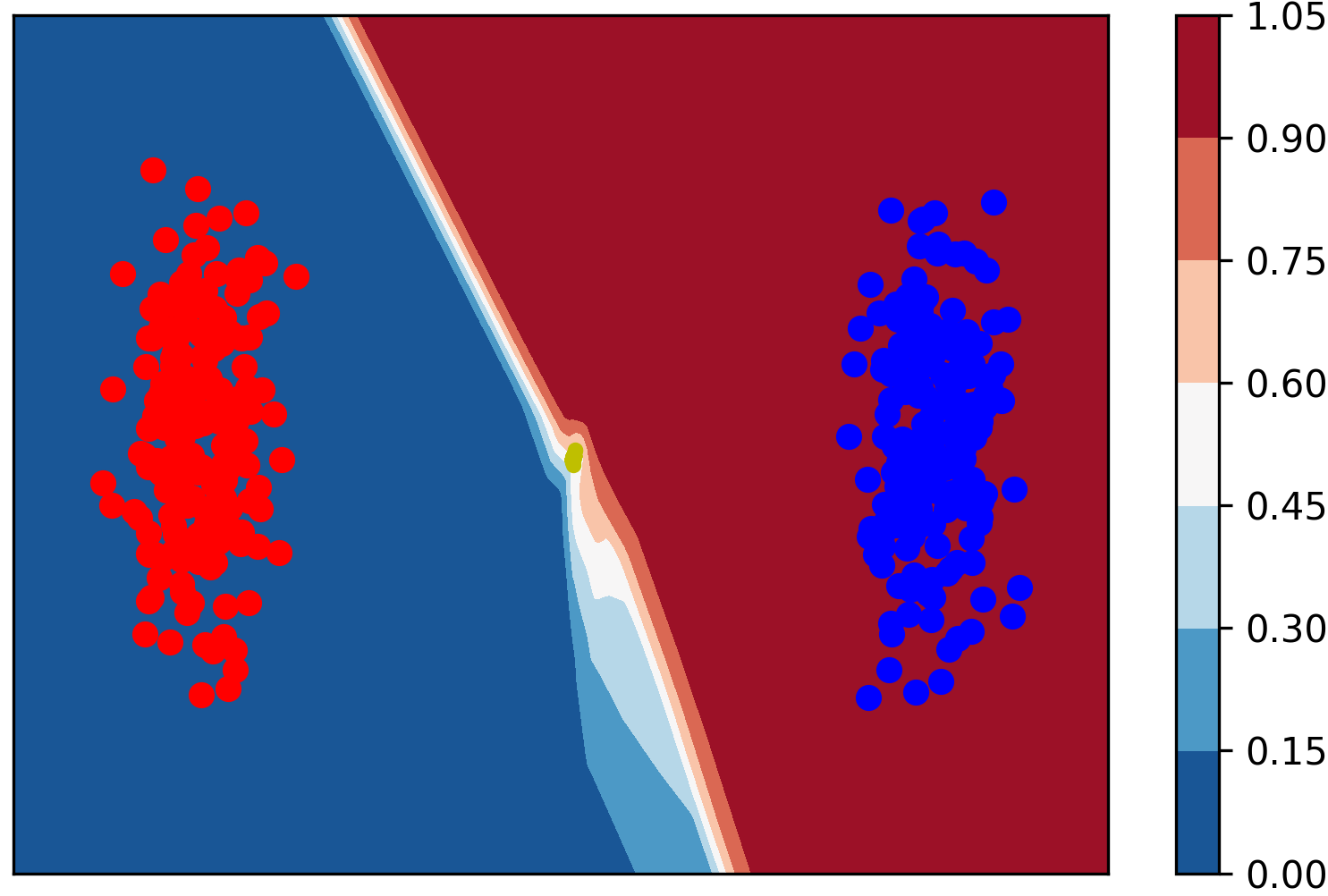} & \includegraphics[width=0.32\linewidth]{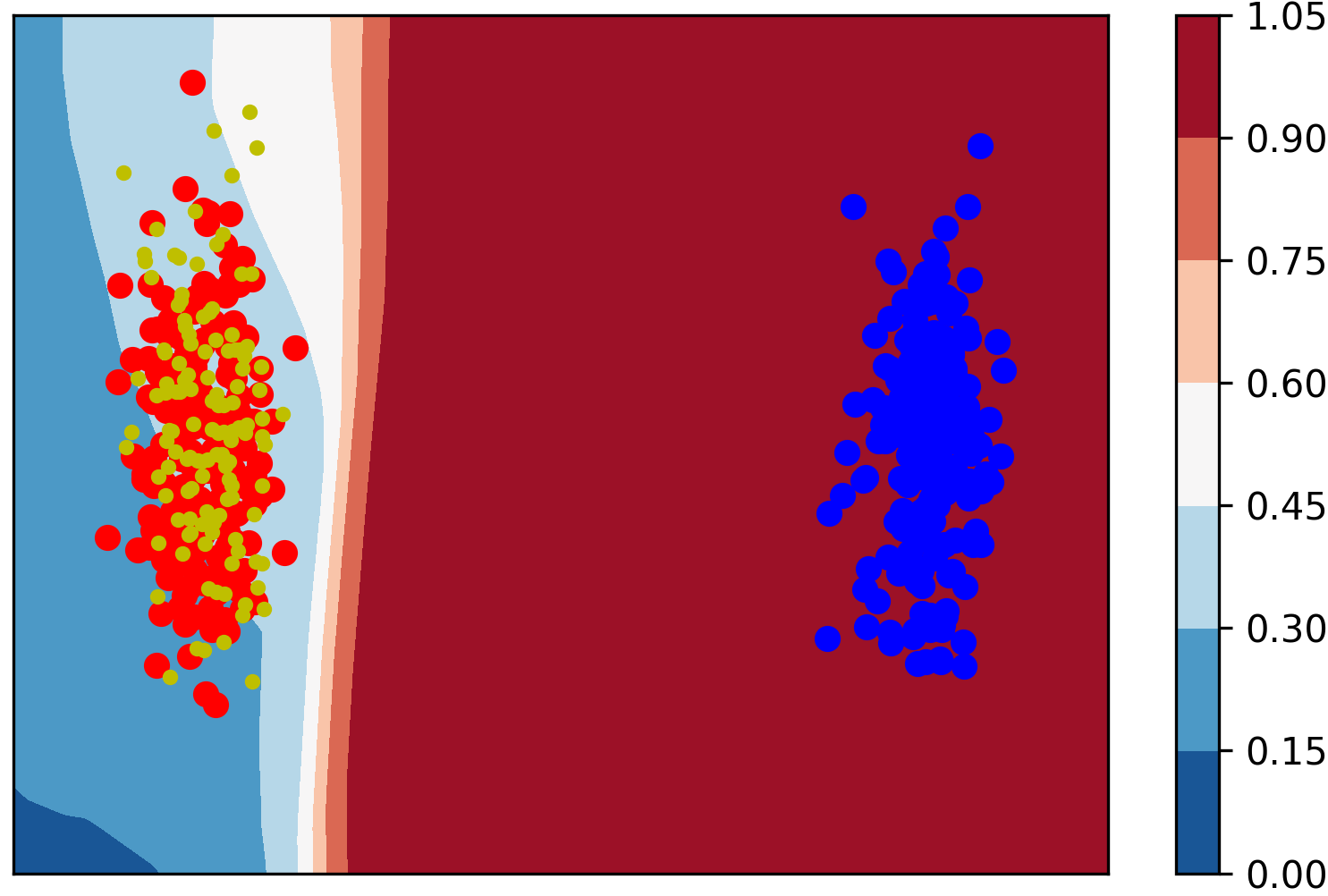} & \includegraphics[width=0.32\linewidth]{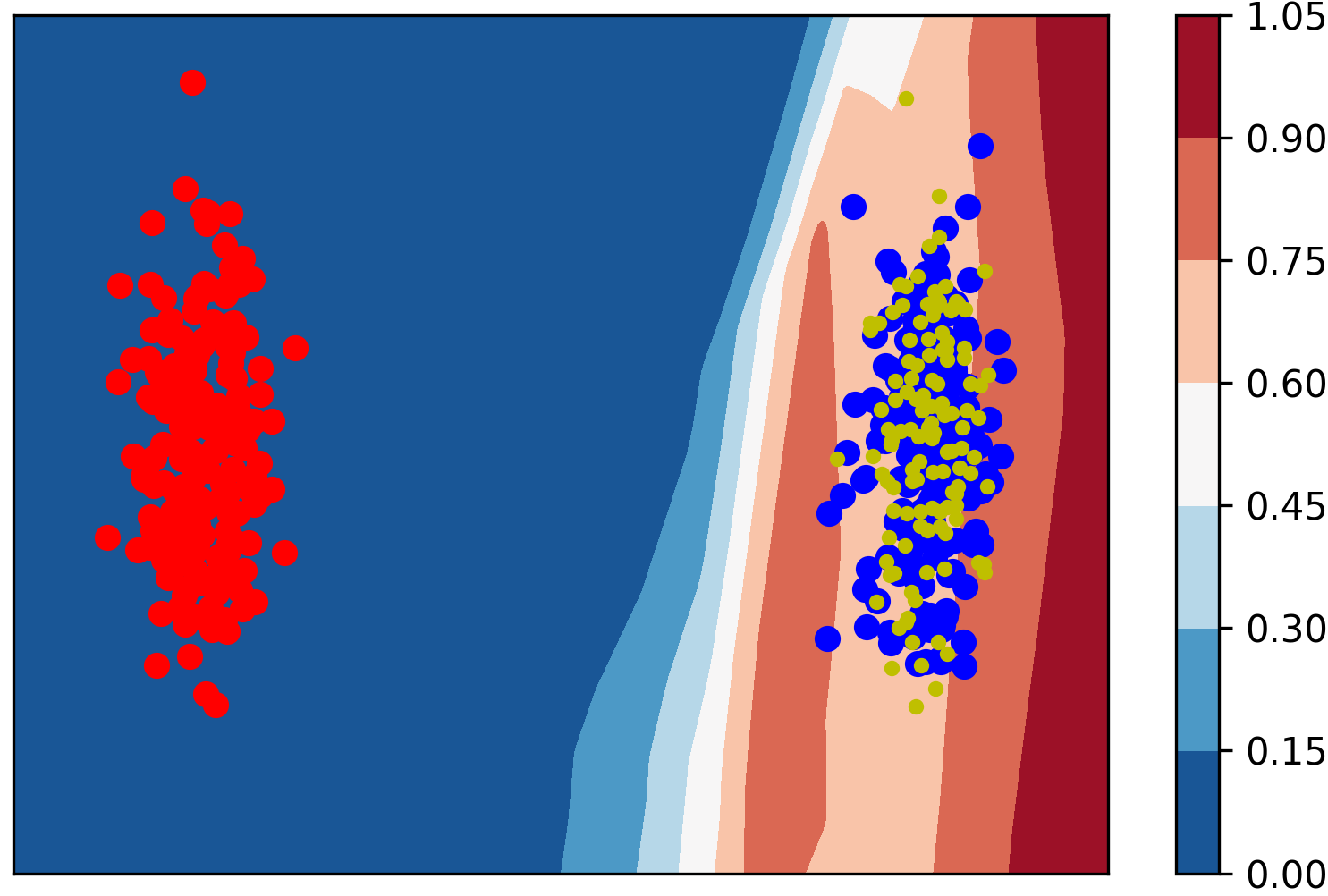} \\
(a) $D_{\text{attr}}$ & (b) $D_{\text{pos}}$ & (c) $D_{\text{neg}}$
\end{tabular}
\caption{Two-Gaussian toy dataset. We translate points in one Gaussian to the other Gaussian and produce confidence maps by (a) a conventional classifier and a bi-directional classifier, where (b) is $D_{\text{pos}}$ aiming for positive-to-negative translation and (c) is $D_{\text{neg}}$ aiming for negative-to-positive translation. Red, blue, and yellow dots represent points in two Gaussian and translated points, respectively. Note that optimal generation often happens near the probability 0.5. Thus, an accurate boundary could benefit translation.}\label{fig:boundary}
\end{figure}

\myparagraph{Partial Inversions using $D_{\text{attr}}$}
Another key issue in performing attribute manipulation in our setting using $D_{\text{attr}}$ arises from high-confidence predictions in low-density data regions.
We find such discriminator feedback encourages the generator to sample partially inverted images in the low-density region where although the discriminator is correctly fooled, the presence/absence of the attribute is visually ambiguous due to proximity to the decision boundary.
For instance, as shown in Fig. \ref{fig:conf_example}, the generated images $x_2$ and $x_3$ (close to vertical gray decision boundary) leads to high-confidence presence/absence predictions (of attribute `blonde'), although both images are visually indistinguishable. 

\myparagraph{Bi-directional Discriminator}
Our core idea to tackle the partial inversion generation problem is to encourage tighter decision boundaries around the positive and negative classes.
We achieve this using a bi-directional discriminator composed of two attribute classifiers: $D_{\text{pos}}$ (to identify positive$\rightarrow$negative image translations) and $D_{\text{neg}}$ (negative$\rightarrow$positive).
In Fig. \ref{fig:conf_example} this is illustrated by the green and orange vertical lines; notice the decision boundaries are closer to high-density regions.

We additionally validate the effectiveness of our bi-directional discriminator over $D_{\text{attr}}$ on a synthetic dataset composed of two Gaussians.
As shown in Fig. \ref{fig:boundary}, each Gaussian cluster represents positive/negative examples and the goal is to ideally perform axis-aligned bi-directional translation (red$\leftrightarrows$blue) from one cluster to another.
In Fig. \ref{fig:boundary}(a), we see that using $D_{\text{attr}}$ leads to translated samples (yellow points) to collapse in a low-density region near the decision boundary.
Our discriminator (Fig. \ref{fig:boundary}(b, c)) produces reasonable translations (where translated yellow points are now in high-density regions) aided by tighter decision boundaries.

We implement bi-directional discriminator using two classifiers $D_{\text{pos}}$ and $D_{\text{neg}}$ (also illustrated in Fig. \ref{fig:model}), where $D_{\text{pos}}$ judges positive-to-negative attribute inversion and $D_{\text{neg}}$ judges negative-to-positive. 
We extend standard binary cross entropy loss ($\mathcal{L}_{\text{bce}}$) to a bi-directional loss ($\mathcal{L}_{\text{bi}}$) to satisfy our constraint:
\begin{equation}
\begin{aligned}
\mathcal{L}_{\text{bi}}(x, y^{\text{org}}, y^{\text{tar}}) =& y^{\text{org}} \mathcal{L}_{\text{bce}}(D_\text{pos}, x, y^\text{tar}) \\ &+ (1-y^\text{org})\mathcal{L}_\text{bce}(D_{\text{neg}}, x, y^\text{tar}),
\end{aligned}
\end{equation}
where $x$ is the input image, $y^{org}$ and $y^{tar}$ denote the original and target labels, and $D_\cdot$ is the classifier used to compute loss function.
The overall objective function is realized as follows
\begin{equation}
    \max_{G}\max_{D}\mathcal{L}_\text{bi}(x, y, y) + \mathcal{L}_\text{bi}(\bar{x}, y, \bar{y}),
\end{equation} where $\bar{x}$ is translated images produced by $G$ along with the target labels $\bar{y}$.
Thus, the bi-directional classifier can be smoothly applied to perform attribute inversion. 
In Figure~\ref{fig:boundary}(b, c) we observe that the bi-directional classifier can provide tighter decision boundaries for both directions and perform effective axis-aligned translations.
Empirically, we find that the performance is sufficiently satisfactory when the discriminator and the bi-path classifier share the same feature extractor and only differ in the last few layers.

\subsection{Learning to Invert Attributes}
\label{ssec:learning}
The proposed framework for the first stage of our approach is trained to minimize a weighted sum of loss functions which regularize the model to achieve our goal discussed in Section~\ref{ssec:method_overview}:
\begin{equation}
\label{eq:overall_loss}
\mathcal{L}_G = \mathcal{L}_\text{rec} + \mathcal{L}_\text{cclf} + \mathcal{L}_\text{bi} +\mathcal{L}_\text{adv} +\lambda_{1}\mathcal{L}_\text{util} +\lambda_{2}\mathcal{L}_\text{reg},
\end{equation}
where $\lambda_{1}$ and $\lambda_{2}$ together controls the trade-off between utility and privacy. We introduce each loss functions in detail over the following paragraphs.

\myparagraph{Reconstruction Loss} 
Given input images $x$, we train the encoder $E$ to produce disentangled representations $(u, c) = E(x)$ to characterize attribute-independent visual features $u$ and disentangled attribute representation $c$.
We adopt an $L_1$ loss to enforce the reconstructed images $\hat{x} = G(u, c)$ to resemble input images $x$:
\begin{equation}
\label{eq:rec_loss}
    \mathcal{L}_\text{rec}(\hat{x}, x) = \left\| \hat{x} - x \right\|_{1}. 
\end{equation}
With this process, we ensure the information contained in images are well-preserved.

\myparagraph{Code Classification Loss} To encode the attribute information into $c$, a mean square loss is imposed on $c$.
Therefore, each element $c_i$ in $c$ represents an binary attribute of $x$.
\begin{equation}
\label{eq:code_clf_loss}
    \mathcal{L}_\text{cclf}(c, y) = \left\| c - y \right\|_{2}^{2},
\end{equation}
where $y$ is the ground truth of sensitive attributes.

Apart from image reconstruction, the decoder generates attribute-inverted images $\Bar{x}=G(u, \Bar{c})$ given the modified sensitive code $\Bar{c}$ along with the non-sensitive code $u$. In particular, we first create $\Bar{c}$ by replacing certain elements of $c$ with binary variables $s \in \{0, 1\}^{N_s} \sim \text{Cat}(K=2, p=0.5)$ and define the modified label $\Bar{y}$ as follows.
\begin{equation}
\label{eq:modified_y}
    \bar{y}_i =
\begin{cases}
    s & \text{if } c^\prime_i \neq c_i\\
    y_i,              & \text{otherwise}
\end{cases}
\end{equation}
The number of inverted attributes $N_s$ is determined by $\text{Cat}(K=\frac{N_A}{2}, p=\frac{2}{N_A})$, which provides the flexibility that the model can invert multiple attributes simultaneously. Note that during the test time, every element $c_i$ can be arbitrary assigned to either $0$ or $1$.

\myparagraph{Bi-Directional Attribute Loss} 
The attribute-inverted images $\Bar{x}$ are required to fool the attribute classifiers in an adversarial manner. As motivated in Section~\ref{ssec:confused_generator}, we apply the bi-directional loss to avoid the partial attribute inversion problem. 
For the generator, we force the generated images to align with $\bar{y}$.
\begin{equation}
\mathcal{L}_\text{bi}(\bar{x}, y, \bar{y}) = y \mathcal{L}_\text{bce}(D_\text{pos}, \bar{x}, \bar{y}) + (1-y)\mathcal{L}_\text{bce}(D_\text{neg}, \bar{x}, \bar{y})
\end{equation}
The above contrasts classifiers which are typically trained to recognize the original attributes, where $m$ masks out positions that $c_{i}$ is not edited:
\begin{equation}
\mathcal{L}_\text{attr}(\bar{x}, y) = m \cdot [y \mathcal{L}_\text{bce}(D_\text{pos}, \bar{x}, y) + (1-y)\mathcal{L}_\text{bce}(D_\text{neg}, \bar{x}, y)],
\end{equation}

\myparagraph{Image Adversarial Loss} 
In addition to fooling the attribute classifiers, we also impose image adversarial loss to encourage the realistic image generation. The intuition is that, without the constraint, the model could generate adversarial examples to fool the attribute classifiers, which violates our motivation.
\begin{equation}
\label{eq:img_adv_loss}
    \mathcal{L}_\text{adv}(\Bar{x}, x) = \log D_\text{img}(x) + \log (1-D_\text{img}(\Bar{x}))
\end{equation}

\myparagraph{Content Regularization Loss}
The attribute-inverted images should resemble the original images although some of attributes are modified. We additionally introduce cycle-consistent reconstruction to the model, encouraging the model to preserve the major content of the original images. 
We introduce the notion of margin to form hinge loss, which balances the tradeoff between privacy and content distortion.
\begin{equation}
    \mathcal{L}_\text{reg}(\Bar{x}, x) = \max (\left\| E(\Bar{x}) - E(x) \right\|_{1} - \delta_1, 0 ),
\end{equation}
where $\delta_1$ indicates tolerance of content distortion.

\myparagraph{Utility Loss} 
In addition to preserving content, non-target attributes, namely those with unchanged $c_i$, also need to be preserved. We ensure it by classical binary cross entropy loss. Similarly, the loss function is controlled by the margins. Note that we impose the loss function on both reconstructed and sanitized images to facilitate the learning. 
\begin{equation}
    \mathcal{L}_\text{util} = \max (\mathcal{L}_\text{bi}(\bar{x}, y)- \delta_2, 0 ),
	+ \max (\mathcal{L}_\text{bi}(\hat{x}, y)- \delta_3, 0 ),
\end{equation}
where $\delta_2$ and $\delta_3$ indicate tolerance of attribute distortion for the sanitized and reconstructed images, respectively. The margin $\delta_3$ is often set to be zero since attributes of reconstructed images are unchanged.

\subsection{Attribute Obfuscation (Stage II)}
\label{ssec:attr_obfu}
With our model (Fig. \ref{fig:model}) trained to minimize loss terms (Eq. \ref{eq:overall_loss}), we are equipped to \textit{invert} attributes i.e., perform bi-directional translations by manipulating presence and absence of targeted attributes in an input. 
Now, we extend the approach to \textit{obfuscate} the image i.e., introduce \textit{uncertainty} over targeted attributes.
To achieve obfuscation, given an input image $x$, we first generate its complement $\bar{x}$ by inverting the presence of the target attribute.
We then generate the obfuscated image $x'$ as a linear interpolation between $x$ and $\bar{x}$:
\begin{equation}
    x' = I_\text{mix}(x, \bar{x}, \lambda) = \lambda x + (1 - \lambda) \bar{x},
\end{equation}
where the mixing coefficient $\lambda \in [0, 1]^{H \times W}$ is generated to maximize the prediction uncertainty with respect to the target attribute.

We train a network $f$ to predict image-specific mixing coefficients $\lambda$:
\begin{equation}
\label{label:eq:lambda_predictor}
\lambda = f(x, \Bar{x}, c, \Bar{c}),
\end{equation}
where $x$ is the input image, $\Bar{x}$ is the attribute-inverted image, $c$ is the sensitive code, and $\Bar{c}$ is the modified sensitive code. 
We model $f$ using 5 residual blocks followed by a 1$\times$1 filter.
The network is trained to produce coefficients that lead to obfuscated images with maximum uncertainty preserving photorealism:
\begin{equation}
\label{eq:loss_second_stage}
    \mathcal{L}_\text{ent}(x^\prime, y^\prime) = \mathcal{L}_\text{adv}(x^\prime) + \mathcal{L}_\text{bi}(x^\prime, y^\prime),
\end{equation}
where $\mathcal{L}_\text{adv}$ encourages $x'$ to be realistic and $\mathcal{L}_\text{bi}$ encourages the interpolated images $x'$ to have maximum entropy with respect to the prediction of both $D_{\text{pos}}$ and $D_{\text{neg}}$ with labels $y'$ (with $p(y_i')$ set to 0.5 for target attribute $i$).

%% file: article/experiments.tex
\section{Experiments}

\subsection{Setup}
\myparagraph{Dataset} 
CelebA is composed of 200K human face images associated with 40 attributes. We choose a subset of 10 disjoint attributes, that is representative of sensitive information. Every input image is center-cropped by 178x178 and then resized to the resolution 128x128. We use 150K images sorted by indices as training data and form a balanced dataset for evaluation from the remaining 53K images.

\myparagraph{Modeling the Adversary} 
To fairly compare our method to prior works, we train a ResNet-18~\cite{he2016deep} classifier on the same training data, which acts as an adversary attempting to infer privacy attributes from images.
The adversary ResNet classifier used during evaluation is:
(i) trained independently to our method; and 
(ii) significantly more complex than the attribute discriminators in our method.
Furthermore, as this attribute classifier achieves near-perfect attribute prediction accuracy, we argue it models a strong unseen adversary to evaluate obfuscation techniques.


\subsection{Evaluation Metrics}
\label{ssec:metrics}

We now present evaluation metrics for both stages of our approach: stage 1 (which \textit{inverts} the target attribute) and stage 2 (which \textit{obfuscates} i.e., maximizes uncertainty of the target attribute).

\myparagraph{Evaluating Attribute Inversions}
We consider the following metrics:
(i) True Positive Rate (TPR = TP / (TP + FN)): to evaluate how well we are able to `remove' the target attribute;
(ii) True Negative Rate (TNR = TN / (TN + FP)): to evaluate effectiveness of `adding' the target attribute; and
(iii) Accuracy.
Note that in all these cases, low scores imply effective inversions.

\myparagraph{Evaluating Attribute Obfuscation}
We evaluate the uncertainty performance by comparing the posterior probabilities (using a held-out classifier $F$) before ($y = F(x)$) and after ($\bar{y} = F(\bar{x})$) image obfuscation.
Specifically, we consider Shannon entropy $H(y)$ to measure the uncertainty (maximum entropy = 1) and additionally observe the confidence of the prediction $y$ (maximum uncertainty at $p(y_i)$ = 0.5) to evaluate attribute obfuscation.

\begin{table}[t]
\footnotesize
\setlength\tabcolsep{0.25em}
\centering
\begin{tabular}{@{}clrrrrrrrr@{}}
\toprule
 & \multicolumn{1}{c}{} & \multicolumn{1}{c}{\begin{tabular}[c]{@{}c@{}}\tiny{Blonde}\\ \tiny{hair}\end{tabular}} & \multicolumn{1}{c}{\tiny{Eyeglass}} & \multicolumn{1}{c}{\begin{tabular}[c]{@{}c@{}}\tiny{Heavy}\\ \tiny{makeup}\end{tabular}} & \multicolumn{1}{c}{\tiny{Male}} & \begin{tabular}[c]{@{}c@{}}\tiny{No}\\ \tiny{beard}\end{tabular} & \multicolumn{1}{c}{\begin{tabular}[c]{@{}c@{}}\tiny{Wavy}\\ \tiny{hair}\end{tabular}} & \multicolumn{1}{c}{\begin{tabular}[c]{@{}c@{}}\tiny{Wearing}\\ \tiny{lipstick}\end{tabular}} & \multicolumn{1}{c}{\tiny{Young}} \\ \midrule
\multirow{4}{*}{\rot{TPR (\%)}} & \texttt{Real} & 100.0 & 100.0 & 100.0 & 100.0 & 100.0 & 100.0 & 100.0 & 100.0 \\
 & \texttt{IcGAN}~\cite{perarnau2016invertible} & 61.6 & 12.1 & 39.6 & 43.8 & 97.5 & 53.1 & 46.3 & 75.6 \\
 & \texttt{StarGAN}~\cite{choi2018stargan} & 10.2 & 2.5 & 65.1 & 41.9 & 51.3 & 36.6 & 65.5 & \textbf{22.8} \\
 & \texttt{Ours} & \textbf{6.0} & \textbf{1.7} & \textbf{21.0} & \textbf{4.3} & \textbf{10.8} & \textbf{28.6} & \textbf{13.7} & 30.1 \\ \cmidrule(l){2-10} 
\multirow{4}{*}{\rot{TNR (\%)}} & \texttt{Real} & 100.0 & 100.0 & 100.0 & 100.0 & 100.0 & 100.0 & 100.0 & 100.0 \\
 & \texttt{IcGAN}~\cite{perarnau2016invertible} & 31.2 & 84.3 & 85.6 & 63.7 & \textbf{5.69} & 65.2 & 79.8 & 37.7 \\
 & \texttt{StarGAN}~\cite{choi2018stargan} & 22.7 & 7.5 & 74.5 & 36.7 & 17.2 & 41.4 & 80.3 & \textbf{25.0} \\
 & \texttt{Ours} & \textbf{14.3} & \textbf{5.5} & \textbf{8.0} & \textbf{3.3} & 15.3 & \textbf{28.6} & \textbf{9.3} & 29.3 \\ \cmidrule(l){2-10} 
\multirow{4}{*}{\rot{Acc (\%)}} & \texttt{Real} & 100.0 & 100.0 & 100.0 & 100.0 & 100.0 & 100.0 & 100.0 & 100.0 \\
 & \texttt{IcGAN}~\cite{perarnau2016invertible} & 35.6 & 79.5 & 67.4 & 55.6 & 82.6 & 61.3 & 63.7 & 66.5 \\
 & \texttt{StarGAN}~\cite{choi2018stargan} & 16.4 & 5.0 & 69.8 & 39.3 & 34.3 & 39.0 & 72.9 & \textbf{23.9} \\
 & \texttt{Ours} & \textbf{10.2} & \textbf{3.6} & \textbf{14.5} & \textbf{3.7} & \textbf{13.1} & \textbf{28.6} & \textbf{11.5} & 29.7 \\ \bottomrule
\end{tabular}
\vspace{1.0em}
\caption{Quantitative results for attribute inversion. Lower (adversary) scores are better.}
\label{table:stat_eval}
\end{table}

\begin{figure*}[tbp]
\centering
\includegraphics[width=0.8\linewidth]{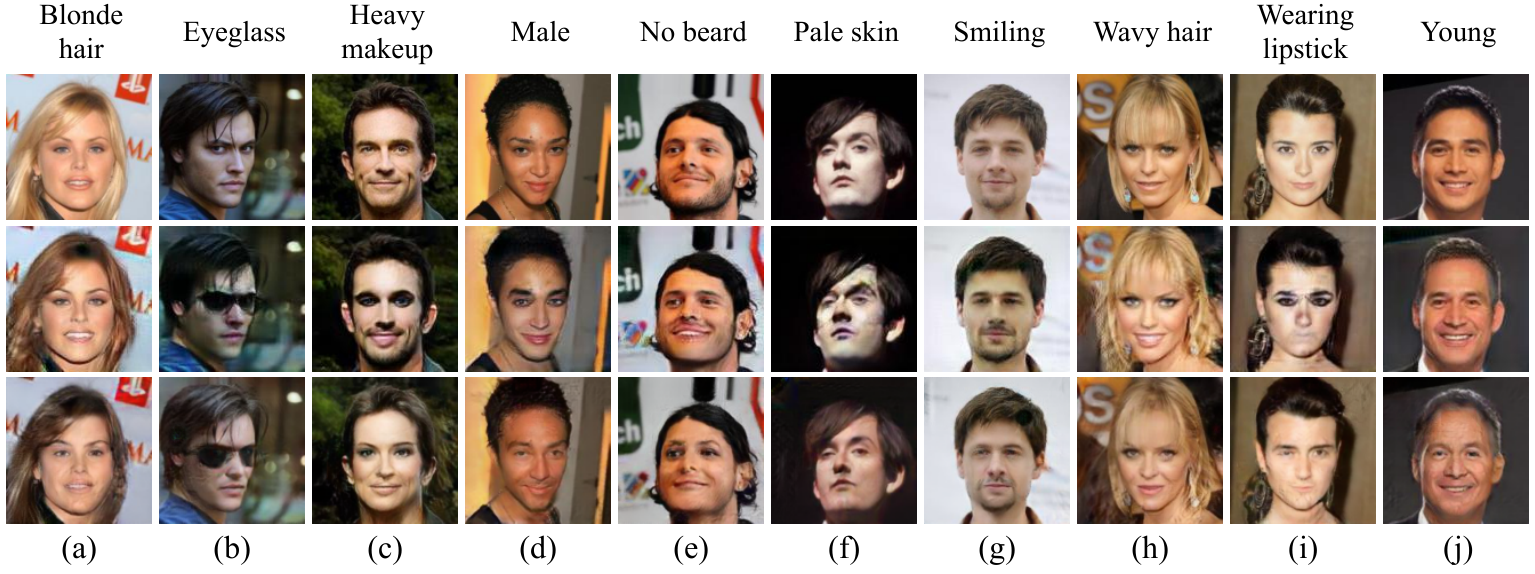}
\caption{Attribute inversion qualitative results. (top) input images; (middle) attribute-inverted images by StarGAN; (bottom) our method.}
\label{fig:one-attr-flip}
\vspace{-0.5em}
\end{figure*}

\subsection{Evaluation against Unseen Adversary's Attack}
\label{ssec:attr_flip}
We verify that our approach to invert attribute presence in images can better conceal inferences over sensitive attributes and generalize to an unseen adversary as compared to typical GAN-based models. In particular, we consider two baselines: IcGAN~\cite{perarnau2016invertible} trains an encoder to map images to input space of a pretrained conditional GAN. By modifying condition vectors encoded from images, IcGAN can manipulate attributes of corresponding inputs. On the other hand, StarGAN~\cite{choi2018stargan} combines conditional GANs with cycle consistency loss to ensure image contents. Note that attributes classifiers in both methods are designed to solely fit the real images. To evaluate robustness against unseen adversaries, we first sanitize images in the testing set for 10 attributes, respectively, and then obtain the prediction accuracy from the held-out ResNet-18. Note that the best case is to minimize accuracy because the original attributes are completely removed. 

\myparagraph{Quantitative Results}
Table~\ref{table:stat_eval} presents the accuracy after sanitization in detail. \textit{Real} denotes the test accuracy of the hold-out classifier on test data. This indicates the generalizability of the classifier to unseen data and ensures the credibility for the following measurement. We observe that the proposed framework reaches the lowest TPR, TNR, and accuracy consistently on most attributes. 
We find the the prior adversarial manipulation methods IcGAN and StarGAN under-perform as they overfit to the attribute-classifier (see Section~\ref{ssec:confused_generator}) during training.
Thus, they do not generalize well to unseen adversaries, especially for challenging attributes such as \textit{Heavy Makeup} and \textit{Male}.

\myparagraph{Qualitative Results}
We visualize samples generated by our method and StarGAN in Figure~\ref{fig:one-attr-flip}. Although both methods generate realistic images, our method can better conceal attributes than StarGAN. For instance, our method completely removes pale skin (Fig. \ref{fig:one-attr-flip}f) while StarGAN only focuses on some specific regions. In addition, our method adds more wrinkles to conceal the age information. In contrast, StarGAN only changes the hair color slightly. Lastly, StarGAN tends to include similar patterns to images as shown in Figure~\ref{fig:one-attr-flip}(c,d), while our method can provide diverse patterns over different attributes. 
From the qualitative results, we find promising results of our approach inverting attributes in images, while being reasonably faithful to the original input image.

\begin{table}[tbp]
\scriptsize
\centering
\begin{tabular}{@{}lrrrrr@{}}
\toprule
 & \multicolumn{5}{c}{Accuracy (\%)} \\ \midrule
 & \multicolumn{1}{c}{\begin{tabular}[c]{@{}c@{}}Blonde \\ Hair\end{tabular}} & \multicolumn{1}{c}{Eyeglass} & \multicolumn{1}{c}{\begin{tabular}[c]{@{}c@{}}Heavy \\ Makeup\end{tabular}} & \multicolumn{1}{c}{Male} & \multicolumn{1}{c}{\begin{tabular}[c]{@{}c@{}}No\\ Beard\end{tabular}} \\ \cmidrule(l){2-6} 
$D_{\text{attr}}$ & 20.09 & 13.11 & 60.68 & 45.23 & 35.83 \\
$D_{\text{attr}}$+AT & 39 & 46.5 & 37.07 & 36.54 & 47.19 \\
Ours & \textbf{10.16} & \textbf{3.61} & \textbf{14.5} & \textbf{3.79} & \textbf{13.07} \\ \midrule
 & \multicolumn{1}{c}{\begin{tabular}[c]{@{}c@{}}Pale \\ Skin\end{tabular}} & \multicolumn{1}{c}{Smiling} & \multicolumn{1}{c}{\begin{tabular}[c]{@{}c@{}}Wavy \\ Hair\end{tabular}} & \multicolumn{1}{c}{\begin{tabular}[c]{@{}c@{}}Wearing \\ lipstick\end{tabular}} & \multicolumn{1}{c}{Young} \\ \cmidrule(l){2-6} 
$D_{\text{attr}}$ & 52.03 & 4.39 & 28.61 & 78.22 & 36.68 \\
$D_{\text{attr}}$+AT & 66.99 & 35.73 & 57.29 & 38.83 & 61.79 \\
Ours & \textbf{35.72} & \textbf{4.18} & \textbf{28.56} & \textbf{11.51} & \textbf{29.67} \\ \bottomrule
\end{tabular}
\vspace{0.5em}
\caption{Ablation study on three models with distinct classifiers.}
\label{table:ablation_study}
\end{table}

\subsection{Ablation Study}
We conduct an ablation study on three models with distinct attribute classifiers to confirm the strength of the proposed bi-directional classifier as introduced in Section~\ref{ssec:confused_generator}. 
In particular, 
(i) \textit{$D_{\text{attr}}$} is equipped with an attribute discriminator solely updated with real data, which is in spirit of traditional image editing methods;
(ii) \textit{$D_{\text{attr}}$+AT} additionally performs adversarial training (AT) by additionally updating $D_{\text{attr}}$ using generated data; and
(iii) \textit{Ours} represents the model equipped with the proposed bi-directional classifier. 
Note that the same encoder-decoder architecture is adopted for all three models.

The accuracy comparison among three models is reported in Table~\ref{table:ablation_study}. We first observe that  \textit{$D_{\text{attr}}$} does not remove attributes thoroughly. 
Ideally, models with adversarially trained classifiers should perform better since it iteratively learns to identify private patterns. 
However, \textit{$D_{\text{attr}}$+AT} performs worse than \textit{$D_{\text{attr}}$}, which confirms the partial attribute inversion problem. 
Lastly, \textit{Ours} reaches the best performance across all attributes. 
The reason is two-fold. 
First, updating the discriminators with generated images makes the model generalize well to unseen classifiers.
Second, the proposed bi-directional classifier further prevents the confusion problem. 

\begin{table}[tbp]
\centering
\footnotesize
\begin{tabularx}{0.8\linewidth}{@{}*{4}{X}@{}}
\toprule
 & $\delta_2=0$ & $\delta_2=0.1$ & $\delta_2=0.2$ \\ \midrule
Privacy & 0.155 & 0.147 & 0.130 \\
Utility & 0.863 & 0.807 & 0.781 \\ \bottomrule
\end{tabularx}
\vspace{0.5em}
\caption{Trade-off between privacy (lower is better) and utility (higher is better).}
\label{table:trade-off}
\end{table}

\subsection{Analysis on Trade-off Parameters}
We present a study of how distinct $\delta_i$ values balance privacy and utility. As discussed in Section~\ref{ssec:learning}, the proposed framework incorporates three parameters $\delta_i$ to control the trade-off. In practice, $\delta_3$ is set to zero as it is related to reconstruction, and $\delta_1$ is often set to be a small number (e.g. 0.05 for L1 norm) since we expect lower distortion. Thus, in this study we mainly focus on the changes of target and non-target attributes when different $\delta_2$ is provided. We denote the accuracy for target attributes by \textit{privacy} and the one for non-target attributes by \textit{utility}.

Ideally, we want to achieve the lowest privacy leakage while maximizing utility.
However, privacy and utility may not be fully independent, leading to a trade-off.
As shown in Table~\ref{table:trade-off}, our model can adjust privacy leakage level by using different $\delta_2$. As desired, if we allow more utility distortion (i.e. larger $\delta_2$), the lower privacy leakage is reached, while the utility distortion is also increased. Users can find suitable parameters based on their applications.


\subsection{Evaluation on Image Quality}
To show that the proposed algorithm can sanitize images without significantly sacrificing image quality, we measure Fréchet Inception Distance (FID) on CelebA for both our algorithm and StarGAN~\cite{choi2018stargan}. We first use each model to generate 50000 images by randomly choosing one target attribute and compute the scores separately on two sets. According to the experiment, our method achieves 9.52 while StarGAN achieves 12.52, which is comparable. This justifies that our method can generate sufficiently high quality images while removing sensitive attributes.

\begin{table}[t]
\footnotesize
\centering
\begin{tabularx}{\linewidth}{@{}l*{8}{R}@{}}
\toprule
& \begin{tabular}[c]{@{}c@{}}\scriptsize{Blonde} \vspace{0.0em}\\  \scriptsize{hair}\end{tabular} & \begin{tabular}[c]{@{}c@{}}\scriptsize{Eye-} \vspace{0.0em}\\ \scriptsize{glasses}\end{tabular} & \begin{tabular}[c]{@{}c@{}}\scriptsize{Heavy} \vspace{0.0em}\\ \scriptsize{makeup}\end{tabular} & \scriptsize{Male} & \begin{tabular}[c]{@{}c@{}}\scriptsize{No} \vspace{0.0em}\\  \scriptsize{beard}\end{tabular} & \begin{tabular}[c]{@{}c@{}}\scriptsize{Wavy} \vspace{0.0em}\\ \scriptsize{hair}\end{tabular} & \begin{tabular}[c]{@{}c@{}}\scriptsize{Wearing} \vspace{0.0em}\\ \scriptsize{lipstick}\end{tabular} & \scriptsize{Young} \\ \midrule
\multicolumn{9}{c}{Entropy (bits)}  \\  \midrule
& \multicolumn{8}{c}{Positive $\rightarrow$ Uncertain}  \\ \cmidrule{2-9}
Real & 0.56 & 0.59 & 0.53 & 0.12 & 0.13 & 0.57 & 0.43 & 0.24 \\
Ours & 0.75 & 0.82 & 0.77 & 0.84 & 0.87 & 0.66 & 0.82 & 0.79 \\
Gain & 0.18 & 0.23 & 0.24 & 0.72 & 0.74 & 0.09 & 0.38 & 0.54 \\ \midrule
& \multicolumn{8}{c}{Negative $\rightarrow$ Uncertain}  \\ \cmidrule{2-9}
Real & 0.06 & 0.04 & 0.08 & 0.23 & 0.47 & 0.16 & 0.08 & 0.46 \\
Ours & 0.83 & 0.89 & 0.80 & 0.86 & 0.66 & 0.72 & 0.81 & 0.64 \\
Gain & 0.78 & 0.84 & 0.72 & 0.63 & 0.19 & 0.56 & 0.73 & 0.18 \\ \midrule
\multicolumn{9}{c}{Probability (\%)}  \\  \midrule
& \multicolumn{8}{c}{Positive $\rightarrow$ Uncertain}  \\ \cmidrule{2-9}
Real & 86.2 & 85.3 & 87.5 & 98.1 & 97.7 & 86.3 & 90.4 & 95.6 \\
Ours & 72.5 & 64.1 & 70.9 & 62.0 & 61.0 & 80.0 & 64.7 & 63.0 \\
Gain & 13.7 & 21.1 & 16.5 & 36.0 & 36.7 & 6.3 & 25.7 & 32.6 \\ \midrule
& \multicolumn{8}{c}{Negative $\rightarrow$ Uncertain}  \\ \cmidrule{2-9}
Real & 0.8 & 0.6 & 1.4 & 4.5 & 10.6 & 2.6 & 1.4 & 10.4 \\
Ours & 44.4 & 47.1 & 37.6 & 38.8 & 22.4 & 38.1 & 38.4 & 20.8 \\
Gain & 43.6 & 46.5 & 36.2 & 34.3 & 11.9 & 35.6 & 37.0 & 10.4 \\ \bottomrule
\end{tabularx}
\vspace{0.5em}
\caption{Quantitative evaluation of attribute obfuscation. Better performance at this task is indicated by higher entropy (maximum = 1 bit) and probability scores approaching 50\% (i.e., chance-level).
`Real' denotes performance of adversary on original non-obfuscated images, and `ours' on obfuscated counterparts. `Gain' denotes the difference between the two.
We evaluate on both positive (input images containing the target attribute) and negative (not containing it).
}
\label{table:uncertainty_eval}
\end{table}

\begin{figure*}[tbp]
\centering
\includegraphics[width=0.8\linewidth]{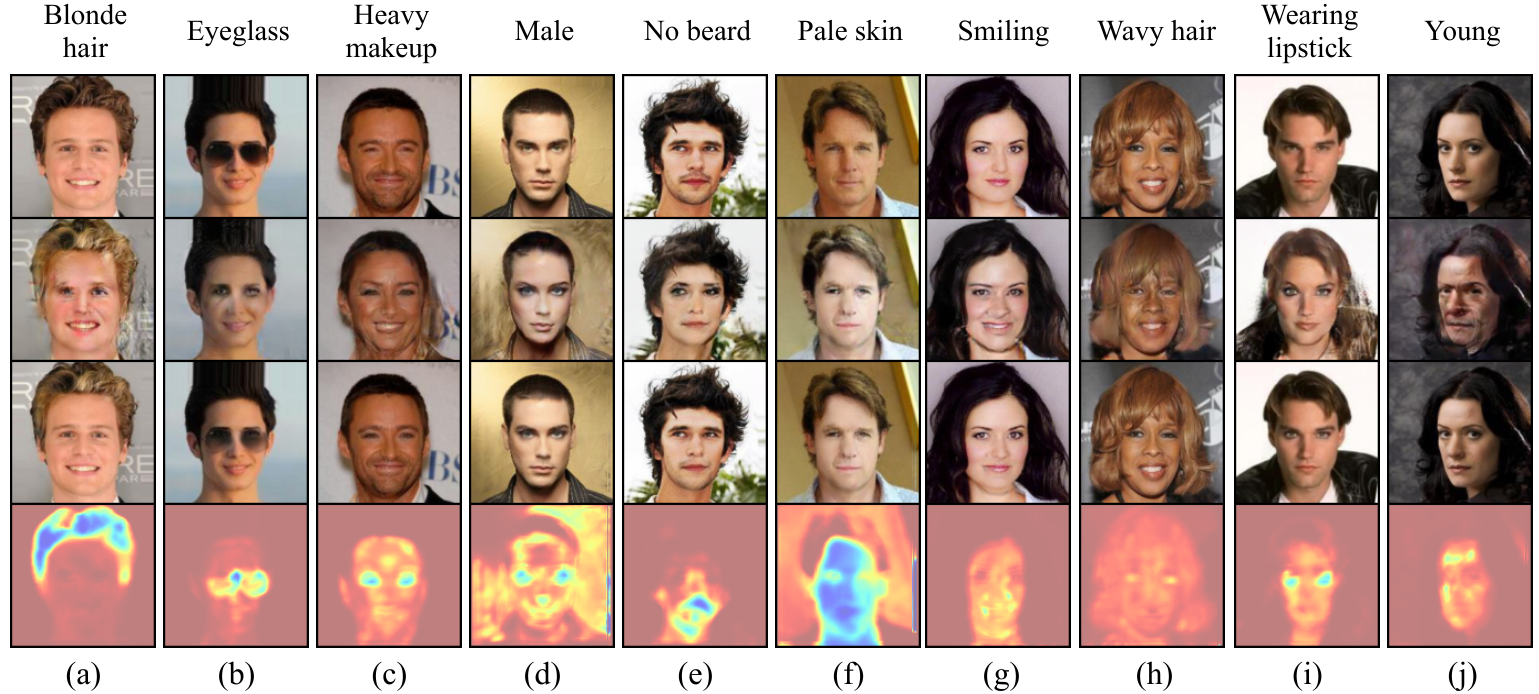}
\caption{Visualization of obfuscated images. From top to bottom, each row presents original images, attribute-inverted images, obfuscated images, and attention maps of $\lambda$. For every column, we choose one attribute as a target.}
\label{fig:obfuscated_imgs}
\vspace{-0.5em}
\end{figure*}

\subsection{Evaluation on Uncertainty}
In the following, we show that the proposed two-stage method can secure privacy information by introducing uncertainty over certain sensitive attributes. Specifically, we consider prediction probability and Shannon entropy to measure privacy leakage. The goal of our method is to minimize leakage, which means ideally, the obfuscated images should have prediction probability 50\% and 1 bit (base 2) entropy over target attributes. Moreover, since vanilla classifiers often suffer from over-confidence problems (i.e. it only outputs either 0\% or 100\%) and thus cause distorted evaluation, we re-train the ResNet-18 classifier with mix-up strategy~\cite{zhang2017mixup}, which mixes two input data and their labels to augment the training data. With the regularization, the model allows ambiguity occurring in predictions and thus prevent over-confident predictions.

In Table~\ref{table:uncertainty_eval}, we report the entropy and prediction probability, for images before (`Real') and after (`Ours') obfuscation, and their corresponding difference (`Gain'). 
We additionally group the results into `(Positive/Negative) $\rightarrow$ Uncertain', where Positive indicates an attribute is present in the input image, and Negative indicating the attribute is absent.
We observe that both entropy and probability are driven toward uncertainty by a large margin, which strongly supports the capability of the proposed two-stage method. 
Interestingly, we find that the Negative $\rightarrow$ Uncertain translation performs better than Positive $\rightarrow$ Uncertain most of the time. This suggests that adding new features to an image is easier than remove information from an image.

We show in Figure~\ref{fig:obfuscated_imgs} that our model can obfuscate sensitive attributes by merging characteristics of original and attribute-inverted images although the images do not necessarily exist in the training data. To name a few, for hair color (Fig.~\ref{fig:obfuscated_imgs} (a)), the model learns to blend blonde into black hair; for male (Fig.~\ref{fig:obfuscated_imgs} (d)), it learns to put on light make-up on the man's face; for pale skin (Fig.~\ref{fig:obfuscated_imgs} (f)), it learns to fuse the face colors. We additionally present interpolation pixel coefficients $\lambda$ in Figure~\ref{fig:obfuscated_imgs}. Surprisingly, the model automatically identifies the regions related to sensitive attributes even though only image-level labels are provided.

%% file: article/conclusion.tex
\section{Conclusion}
\label{sec:conclusion}

In this paper, we were motivated by providing fine-grained control over private information leakage in images.
Towards this goal, we presented an approach to obfuscate images, where the information w.r.t target privacy attributes is manipulated -- either by inverting the attribute, or maximizing uncertainty over it.
In spite of numerous challenges this setting presents 
(e.g. generating out-of-domain obfuscated data, generalizing to unseen attribute inference attacks), we show that images can be sufficiently altered to either introduce false information, or minimize the information content of an attribute, while maintaining the overall appearance of the original input image.
%


%% file: article/appendix.tex
\noindent \textbf{\LARGE Appendix} \\
\setcounter{table}{0}
\renewcommand{\thetable}{A\arabic{table}}
\setcounter{figure}{0}
\renewcommand{\thefigure}{A\arabic{figure}}


\section{Scatter plots}

\begin{figure*}[b]
    \centering
    \subfigure[Obfuscating images \textit{containing} attribute $y$ originally]{\includegraphics[width=1.0\linewidth]{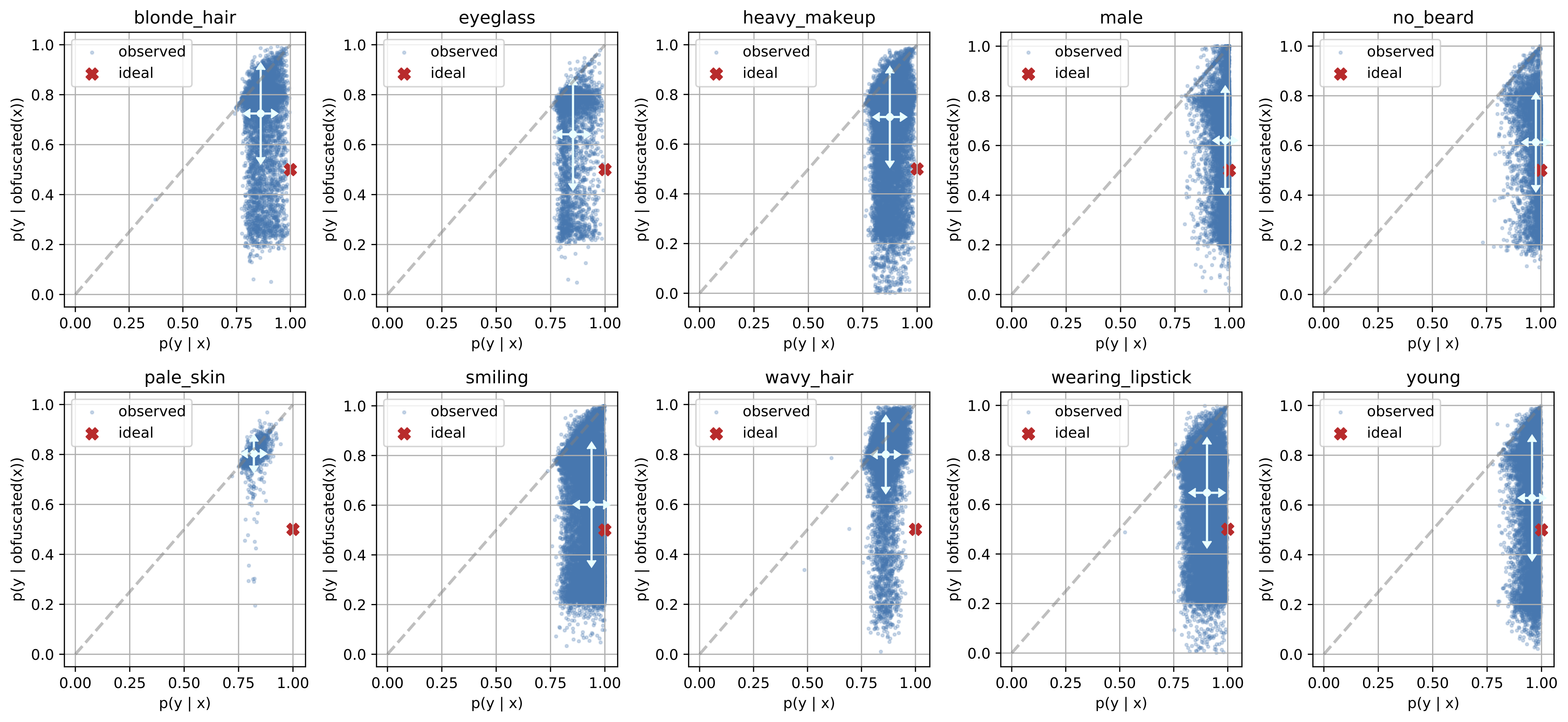}}
    \subfigure[Obfuscating images \textit{not containing} attribute $y$]{\includegraphics[width=1.0\linewidth]{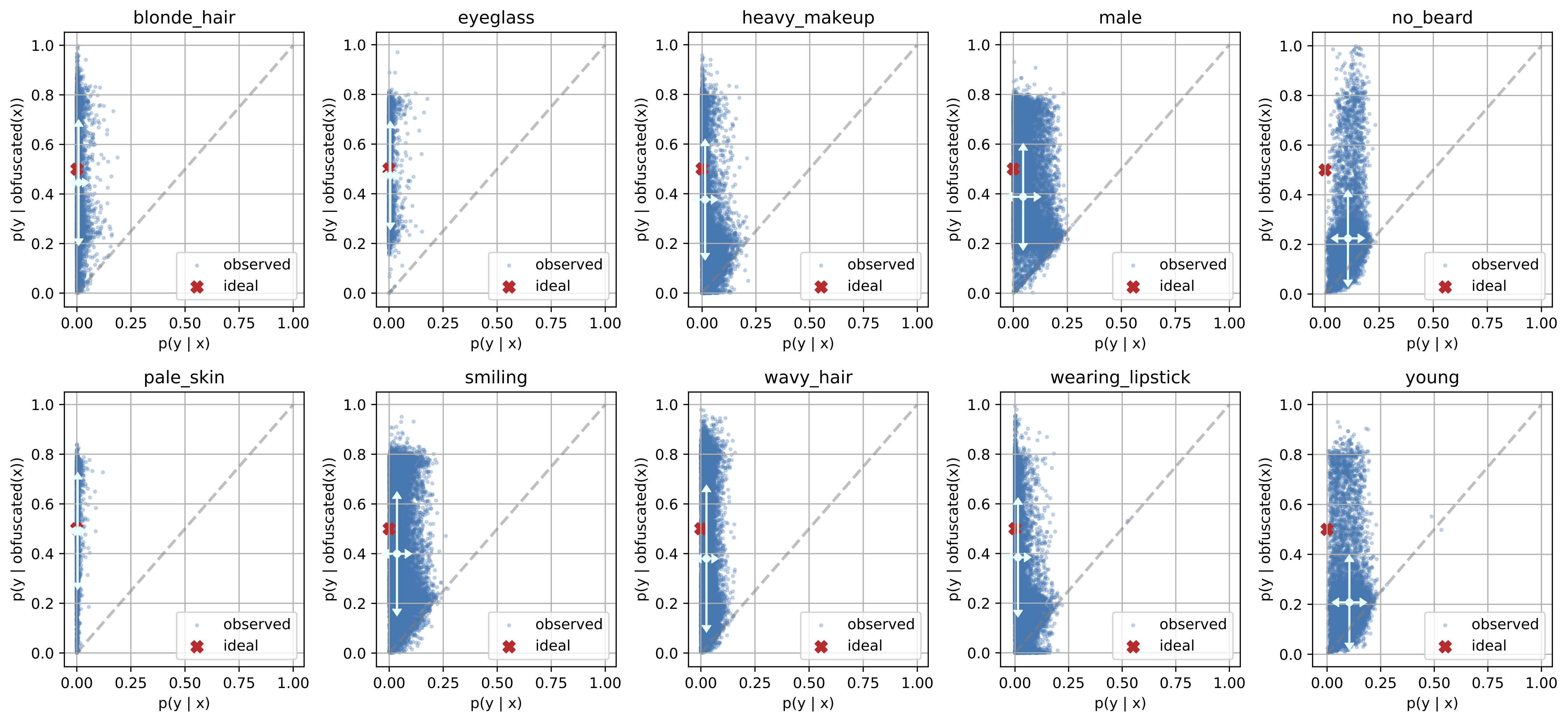}}
    \caption{Each point in the scatter plot represents posterior probabilities before ($x$-axis) and after ($y$-axis) obfuscating attribute $y_i$ (see plot title) in image $x_i$. The probabilities are obtained using a held-out classifier. The light blue marker with bars denotes the mean and standard deviation of the distribution.}
\label{fig:scatter}
\end{figure*}

In Figure \ref{fig:scatter}, we present the predicted posteriors before and after obfuscation, with each plot denoting obfuscation of a particular attribute (indicated by title of each plot). 
We further cluster the obfuscation results into images that originally contained the attribute targeted for obfuscation (see Fig. \ref{fig:scatter}(a), and which did not originally contain the attribute (see \ref{fig:scatter}(b)).

Across all plots, we desire the predictions for the obfuscated image (i.e., `ideal') to collapse around $p(y|\text{obfuscated}(x)) = 0.5$, which has the maximum entropy of 1 bit.
From Figure \ref{fig:scatter}, across all attributes, we observe strong performance: the attribute predictions of obfuscated images move towards the maximum entropy region.

\section{Histograms of Predictions}

We now visualize the distribution of inferences of attribute $y$, for various choices of attributes and image manipulation strategies.
Specifically, we present statistics of binary attribute predictions $p(y | x)$, where:
(i) $x$ = original: i.e., prediction on the original image;
(ii) $x$ = inverted: the presence/absence of attribute $y$ in the image is flipped; and
(iii) $x$ = obfuscated: uncertainty over attribute $y$ is maximized.

\begin{figure*}[t]
    \centering
    \subfigure[Predicted Probabilities]{\includegraphics[width=0.4\linewidth]{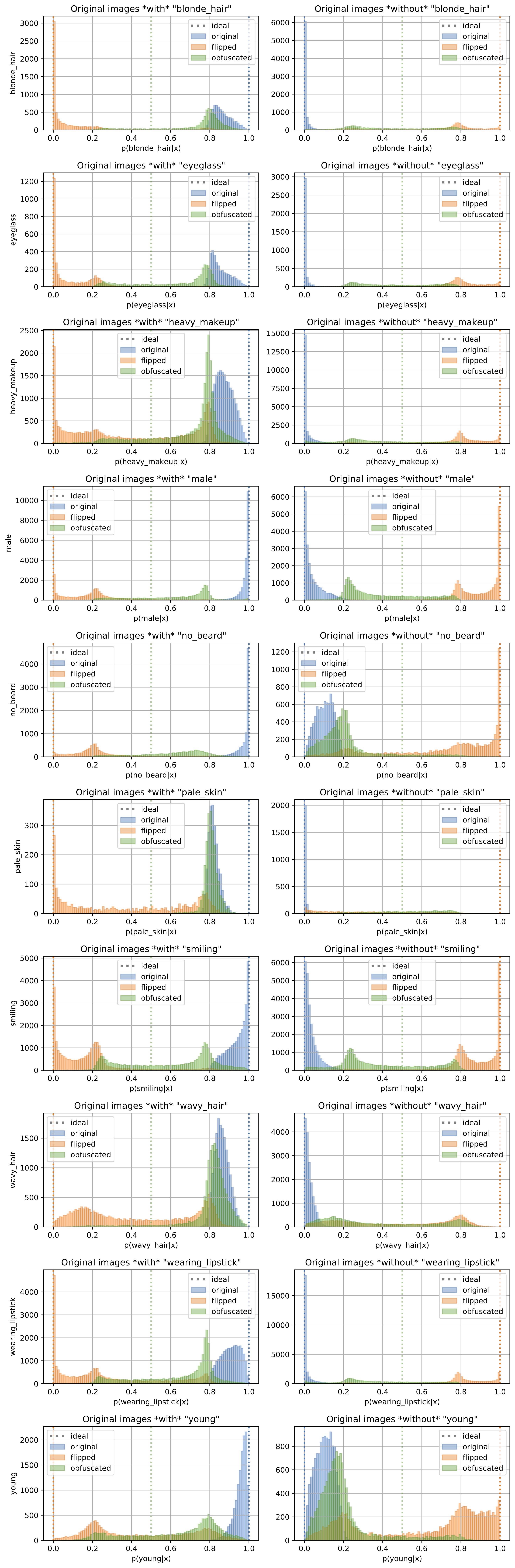}}
    \subfigure[Entropies of Predictions]{\includegraphics[width=0.4\linewidth]{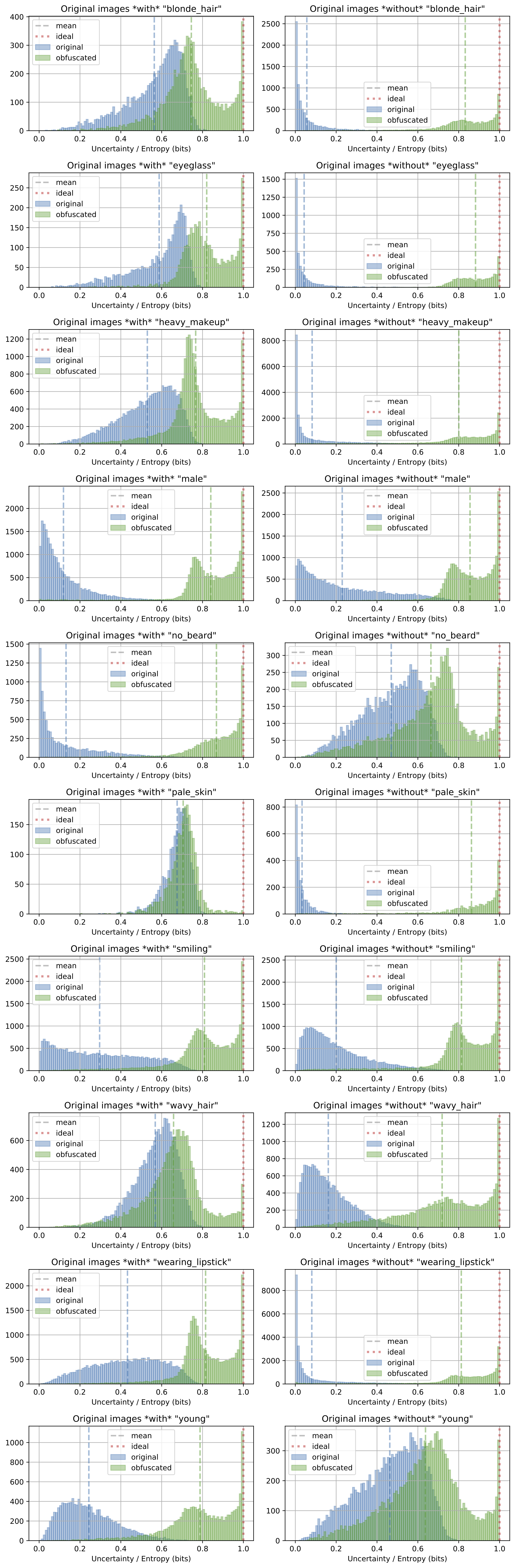}}
    \caption{Histogram of predictions on original and manipulated images.}
\label{fig:histogram}
\end{figure*}

Figure \ref{fig:histogram}(a) presents the histogram over the \textit{probabilities} $p(y | x)$.
We then compute the \textit{entropies} of these predictions and display them alongside in Figure \ref{fig:histogram}(b).
Notice in both cases, we present the results along two columns: left-column presents statistics where the original image's ground-truth label contains attribute $y$, and the right-column where it does not.
From these figures, we observe that our approach is successful in inverting the attribute i.e., the blue distribution (original images) in Fig. \ref{fig:histogram}(a) is reasonably transformed to the yellow distribution (attribute-inverted images) on the opposite side (where $p' = 1 - p$) of the $x$-axis, or as the green distribution (where $p' = 0.5$). 
We find a similar effect on corresponding entropies in Fig. \ref{fig:histogram}(b).
By obfuscating the images, the predictions that were originally confident (hence low entropies in Fig. \ref{fig:histogram}(b) demonstrate a larger uncertainties post-obfuscation (closer to maximum entropy of 1 bit).


\section{Implementation Details}
\label{ssec:method_implement}
We implement our encoder and generator based on the U-net architecture~\cite{ronneberger2015u} and adopt the design of discriminators from StarGAN~\cite{choi2018stargan}. Additionally, we exploit TTUR~\cite{heusel2017gans} and Spectral Normalization~\cite{miyato2018spectral} to stabilize adversarial learning. We use Adam optimizers~\cite{kingma2014adam} with initial learning rate $1\times10^{-4}$ for the autoencoder and $5 \times 10^{-4}$ for discriminators. We train our encoder-decoder model for 300K iterations and the $\lambda$-prediction network $f$ for 100K iterations with batch size 128. Learning rates decrease according to the linear decay strategy. 

\section{Quantitative Evaluation on More Attributes}
Table~\ref{table:stat_eval} and Table~\ref{table:uncertainty_eval} present the same experiments as in Table 1 and Table 4 in the main paper on more attributes.

\begin{table*}[t]
\footnotesize
\setlength\tabcolsep{0.25em}
\centering
\caption{Quantitative results for attribute inversion. Lower (adversary) scores are better.}\label{table:stat_eval}
\begin{tabular}{@{}clrrrrrrrrrr@{}}
\toprule
 & \multicolumn{1}{c}{} & \multicolumn{1}{c}{\begin{tabular}[c]{@{}c@{}}\tiny{Blonde}\\ \tiny{hair}\end{tabular}} & \multicolumn{1}{c}{\tiny{Eyeglass}} & \multicolumn{1}{c}{\begin{tabular}[c]{@{}c@{}}\tiny{Heavy}\\ \tiny{makeup}\end{tabular}} & \multicolumn{1}{c}{\tiny{Male}} & \begin{tabular}[c]{@{}c@{}}\tiny{No}\\ \tiny{beard}\end{tabular} & \multicolumn{1}{c}{\begin{tabular}[c]{@{}c@{}}\tiny{Pale}\\ \tiny{skin}\end{tabular}} & \multicolumn{1}{c}{\tiny{Smiling}} & \multicolumn{1}{c}{\begin{tabular}[c]{@{}c@{}}\tiny{Wavy}\\ \tiny{hair}\end{tabular}} & \multicolumn{1}{c}{\begin{tabular}[c]{@{}c@{}}\tiny{Wearing}\\ \tiny{lipstick}\end{tabular}} & \multicolumn{1}{c}{\tiny{Young}} \\ \midrule
\multirow{4}{*}{\rot{TPR (\%)}} & \texttt{Real} & 100.0 & 100.0 & 100.0 & 100.0 & 100.0 & 100.0 & 100.0 & 100.0 & 100.0 & 100.0 \\
 & \texttt{IcGAN}~\cite{perarnau2016invertible} & 61.6 & 12.1 & 39.6 & 43.8 & 97.5 & 44.0 & 15.6 & 53.1 & 46.3 & 75.6 \\
 & \texttt{StarGAN}~\cite{choi2018stargan} & 10.2 & 2.5 & 65.1 & 41.9 & 51.3 & 26.2 & \textbf{1.9} & 36.6 & 65.5 & \textbf{22.8} \\
 & \texttt{Ours} & \textbf{6.0} & \textbf{1.7} & \textbf{21.0} & \textbf{4.3} & \textbf{10.8} & \textbf{16.9} & 4.4 & \textbf{28.6} & \textbf{13.7} & 30.1 \\ \cmidrule(l){2-12} 
\multirow{4}{*}{\rot{TNR (\%)}} & \texttt{Real} & 100.0 & 100.0 & 100.0 & 100.0 & 100.0 & 100.0 & 100.0 & 100.0 & 100.0 & 100.0 \\
 & \texttt{IcGAN}~\cite{perarnau2016invertible} & 31.2 & 84.3 & 85.6 & 63.7 & \textbf{5.69} & 38.8 & 20.9 & 65.2 & 79.8 & 37.7 \\
 & \texttt{StarGAN}~\cite{choi2018stargan} & 22.7 & 7.5 & 74.5 & 36.7 & 17.2 & \textbf{45.2} & 5.4 & 41.4 & 80.3 & \textbf{25.0} \\
 & \texttt{Ours} & \textbf{14.3} & \textbf{5.5} & \textbf{8.0} & \textbf{3.3} & 15.3 & 54.6 & \textbf{4.0} & \textbf{28.6} & \textbf{9.3} & 29.3 \\ \cmidrule(l){2-12} 
\multirow{4}{*}{\rot{Acc (\%)}} & \texttt{Real} & 100.0 & 100.0 & 100.0 & 100.0 & 100.0 & 100.0 & 100.0 & 100.0 & 100.0 & 100.0 \\
 & \texttt{IcGAN}~\cite{perarnau2016invertible} & 35.6 & 79.5 & 67.4 & 55.6 & 82.6 & 39.0 & 18.3 & 61.3 & 63.7 & 66.5 \\
 & \texttt{StarGAN}~\cite{choi2018stargan} & 16.4 & 5.0 & 69.8 & 39.3 & 34.3 & \textbf{35.7} & \textbf{3.6} & 39.0 & 72.9 & \textbf{23.9} \\
 & \texttt{Ours} & \textbf{10.2} & \textbf{3.6} & \textbf{14.5} & \textbf{3.7} & \textbf{13.1} & \textbf{35.7} & 4.2 & \textbf{28.6} & \textbf{11.5} & 29.7 \\ \bottomrule
\end{tabular}
\end{table*}

\begin{table*}[t]
\small
\centering
\caption{Quantitative evaluation of attribute obfuscation. Better performance at this task is indicated by higher entropy (maximum = 1 bit) and probability scores approaching 50\% (i.e., chance-level).
`Real' denotes performance of adversary on original non-obfuscated images, and `ours' on obfuscated counterparts. `Gain' denotes the difference between the two.
We evaluate on both positive (input images containing the target attribute) and negative (not containing it).
}\label{table:uncertainty_eval}
\begin{tabularx}{0.7\linewidth}{@{}l*{10}{R}@{}}
\toprule
& \begin{tabular}[c]{@{}c@{}}\small{Blonde} \vspace{0.0em}\\  \small{hair}\end{tabular} & \begin{tabular}[c]{@{}c@{}}\small{Eye-} \vspace{0.0em}\\ \small{glasses}\end{tabular} & \begin{tabular}[c]{@{}c@{}}\small{Heavy} \vspace{0.0em}\\ \small{makeup}\end{tabular} & \small{Male} & \begin{tabular}[c]{@{}c@{}}\small{No} \vspace{0.0em}\\  \small{beard}\end{tabular} & \begin{tabular}[c]{@{}c@{}}\small{Pale} \vspace{0.0em}\\ \small{skin}\end{tabular} & \small{Smiling} & \begin{tabular}[c]{@{}c@{}}\small{Wavy} \vspace{0.0em}\\ \small{hair}\end{tabular} & \begin{tabular}[c]{@{}c@{}}\small{Wearing} \vspace{0.0em}\\ \small{lipstick}\end{tabular} & \small{Young} \\ \midrule
\multicolumn{11}{c}{Entropy (bits)}  \\  \midrule
& \multicolumn{10}{c}{Positive $\rightarrow$ Uncertain}  \\ \cmidrule{2-11}
Real & 0.56 & 0.59 & 0.53 & 0.12 & 0.13 & 0.68 & 0.30 & 0.57 & 0.43 & 0.24 \\
Ours & 0.75 & 0.82 & 0.77 & 0.84 & 0.87 & 0.71 & 0.81 & 0.66 & 0.82 & 0.79 \\
Gain & 0.18 & 0.23 & 0.24 & 0.72 & 0.74 & 0.03 & 0.51 & 0.09 & 0.38 & 0.54 \\ \midrule
& \multicolumn{10}{c}{Negative $\rightarrow$ Uncertain}  \\ \cmidrule{2-11}
Real & 0.06 & 0.04 & 0.08 & 0.23 & 0.47 & 0.03 & 0.20 & 0.16 & 0.08 & 0.46 \\
Ours & 0.83 & 0.89 & 0.80 & 0.86 & 0.66 & 0.86 & 0.81 & 0.72 & 0.81 & 0.64 \\
Gain & 0.78 & 0.84 & 0.72 & 0.63 & 0.19 & 0.83 & 0.61 & 0.56 & 0.73 & 0.18 \\ \midrule
\multicolumn{11}{c}{Probability (\%)}  \\  \midrule
& \multicolumn{10}{c}{Positive $\rightarrow$ Uncertain}  \\ \cmidrule{2-11}
Real & 86.2 & 85.3 & 87.5 & 98.1 & 97.7 & 82.0 & 93.9 & 86.3 & 90.4 & 95.6 \\
Ours & 72.5 & 64.1 & 70.9 & 62.0 & 61.0 & 80.4 & 60.2 & 80.0 & 64.7 & 63.0 \\
Gain & 13.7 & 21.1 & 16.5 & 36.0 & 36.7 & 1.7 & 33.6 & 6.3 & 25.7 & 32.6 \\ \midrule
& \multicolumn{10}{c}{Negative $\rightarrow$ Uncertain}  \\ \cmidrule{2-11}
Real & 0.8 & 0.6 & 1.4 & 4.5 & 10.6 & 0.4 & 3.6 & 2.6 & 1.4 & 10.4 \\
Ours & 44.4 & 47.1 & 37.6 & 38.8 & 22.4 & 49.7 & 39.9 & 38.1 & 38.4 & 20.8 \\
Gain & -43.6 & -46.5 & -36.2 & -34.3 & -11.9 & -49.3 & -36.3 & -35.6 & -37.0 & -10.4 \\ \bottomrule
\end{tabularx}
\end{table*}